\documentclass[10pt,twocolumn,letterpaper]{article}

\usepackage{cvpr,times,graphicx,amsmath,amssymb,multirow,array,xcolor}
\usepackage[caption=false]{subfig}
\usepackage{graphicx,amsmath,amssymb}
\definecolor{citecolor}{RGB}{34,139,34}
\usepackage[pagebackref=true,breaklinks=true,letterpaper=true,colorlinks,citecolor=citecolor,bookmarks=false]{hyperref}

\cvprfinalcopy
\def\cvprPaperID{***}

\newcommand{\bd}[1]{\textbf{#1}}
\newcommand{\app}{\raise.17ex\hbox{$\scriptstyle\sim$}}
\newcommand{\cat}[1]{{\emph{#1}}\xspace}
\newcolumntype{x}[1]{>{\centering\arraybackslash}p{#1pt}}
\newcommand{\eqnnm}[2]{\begin{equation}\label{eq:#1}#2\end{equation}\ignorespaces}
\newlength\savewidth\newcommand\shline{\noalign{\global\savewidth\arrayrulewidth
 \global\arrayrulewidth 1pt}\hline\noalign{\global\arrayrulewidth\savewidth}}
\newcommand{\tablestyle}[2]{\setlength{\tabcolsep}{#1}\renewcommand{\arraystretch}{#2}\centering\footnotesize}
\makeatletter\renewcommand\paragraph{\@startsection{paragraph}{4}{\z@}
 {.5em \@plus1ex \@minus.2ex}{-.5em}{\normalfont\normalsize\bfseries}}\makeatother
\newcommand{\paragraphi}[1]{\textit{#1}}

\def\method{InteractNet\xspace}
\newcommand{\symb}[1]{{\small\texttt{#1}}\xspace}

\def\hvo{{\small$\langle$\texttt{human}, \texttt{verb}, \texttt{object}$\rangle$}\xspace}
\def\hv{{\small$\langle$\texttt{human}, \texttt{verb}$\rangle$}\xspace}

\begin{document}

\title{Detecting and Recognizing Human-Object Interactions\vspace{-.5em}}
\author{
 Georgia Gkioxari \quad Ross Girshick \quad Piotr Doll\'ar \quad Kaiming He\vspace{.5em}\\
 Facebook AI Research (FAIR)
}
\maketitle

\begin{abstract}
To understand the visual world, a machine must not only recognize individual object instances but also how they \emph{interact}. Humans are often at the center of such interactions and detecting human-object interactions is an important practical and scientific problem. In this paper, we address the task of detecting \hvo triplets in challenging everyday photos. We propose a novel model that is driven by a \emph{human-centric} approach. Our hypothesis is that the appearance of a person -- their pose, clothing, action -- is a powerful cue for localizing the objects they are interacting with. To exploit this cue, our model learns to predict an action-specific density over target object locations based on the appearance of a detected person. Our model also jointly learns to detect people and objects, and by fusing these predictions it efficiently infers interaction triplets in a clean, jointly trained end-to-end system we call \method. We validate our approach on the recently introduced Verbs in COCO (V-COCO) and HICO-DET datasets, where we show quantitatively compelling results.
\end{abstract}

\section{Introduction}\label{sec:intro}

Visual recognition of individual instances, \eg, detecting objects \cite{Girshick2014,Girshick2015,Ren2015} and estimating human actions/poses \cite{Gkioxari2015a,Wei2016,Cao2017}, has witnessed significant improvements thanks to deep learning visual representations \cite{Krizhevsky2012,Simonyan2015,Szegedy2015,He2016}. However, recognizing individual objects is just a first step for machines to comprehend the visual world. To understand what is \emph{happening} in images, it is necessary to also recognize relationships between individual instances. In this work, we focus on \emph{human-object interactions}.

The task of recognizing human-object interactions \cite{Gupta2009,Yao2010,Delaitre2010,Gupta2015,Chen2014accv} can be represented as detecting \hvo triplets and is of particular interest in applications and in research. From a practical perspective, photos containing people contribute a considerable portion of daily uploads to internet and social networking sites, and thus human-centric understanding has significant demand in practice. From a research perspective, the person category involves a rich set of actions/verbs, most of which are rarely taken by other subjects (\eg, \cat{to talk}, \cat{throw}, \cat{work}). The fine granularity of human actions and their interactions with a wide array of object types presents a new challenge compared to recognition of entry-level object categories.

\begin{figure}[t]
\centering
\includegraphics[width=1.\linewidth]{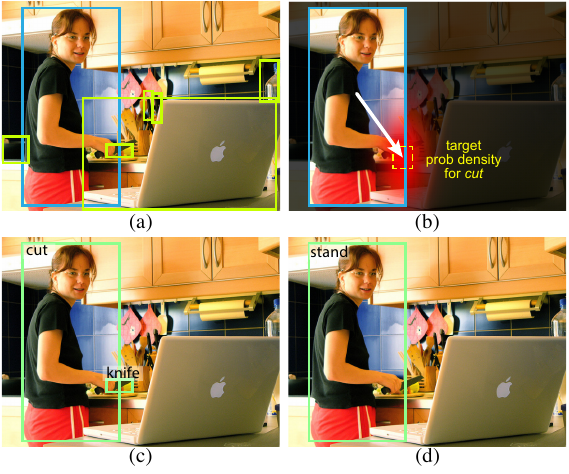}
\caption{\bd{Detecting and recognizing human-object interactions.} \textbf{(a)} There can be many possible objects (green boxes) interacting with a detected person (blue box). \textbf{(b)} Our method estimates an action-type specific \emph{density} over target object locations from the person's appearance, which is represented by features extracted from the detected person's box. \textbf{(c)} A \hvo triplet detected by our method, showing the person box, action (\cat{cut}), and target object box and category (\cat{knife}). \textbf{(d)} Another predicted action (\cat{stand}), noting that a person can simultaneously take multiple actions and an action may not involve any objects.}
\label{fig:teaser}
\vspace{-0.5em}
\end{figure}

In this paper, we present a human-centric model for recognizing human-object interaction. Our central observation is that a person's appearance, which reveals their action and pose, is highly informative for inferring where the target object of the interaction may be located (Figure~\ref{fig:teaser}(b)). The search space for the target object can thus be narrowed by conditioning on this estimation. Although there are often many objects detected (Figure~\ref{fig:teaser}(a)), the inferred target location can help the model to quickly pick the correct object associated with a specific action (Figure~\ref{fig:teaser}(c)).

\begin{figure*}[t]
\centering
\includegraphics[width=1.\linewidth]{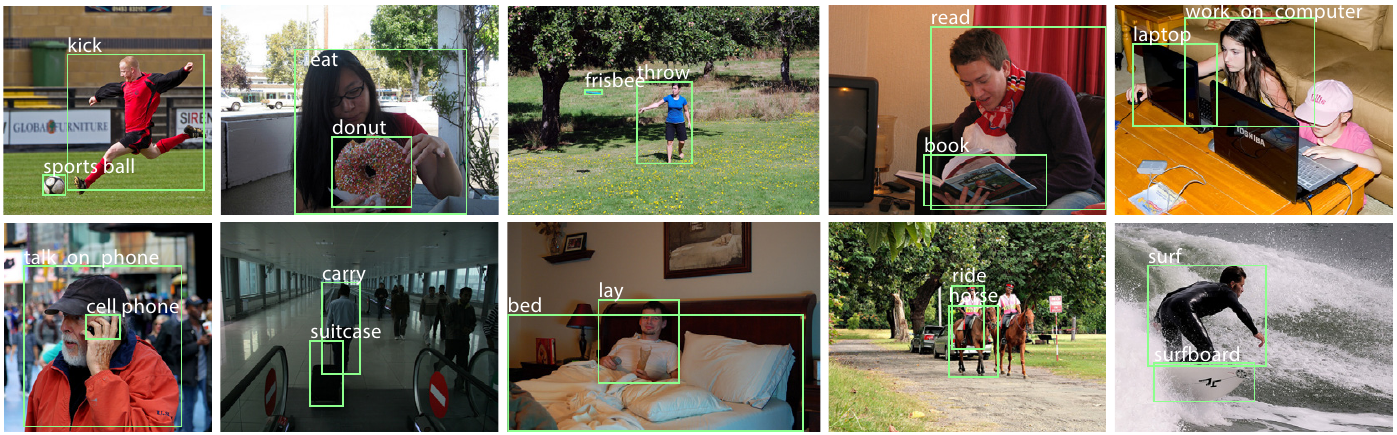}
\caption{Human-object interactions detected by our method. Each image shows one detected \hvo triplet.}
\label{fig:results_simple}
\vspace{-1em}
\end{figure*}

We implement this idea as a \emph{human-centric} recognition branch in the Faster R-CNN framework \cite{Ren2015}. Specifically, on a region of interest (RoI) associated with a person, this branch performs \emph{action} classification and \emph{density estimation} for the action's target object location. The density estimator predicts a 4-d Gaussian distribution, for each action type, that models the likely \emph{relative} position of the target object to the person. The prediction is based purely on the human appearance. This human-centric recognition branch, along with a standard object detection branch \cite{Girshick2015} and a simple pairwise interaction branch (described later), form a multi-task learning system that can be jointly optimized.

We evaluate our method, \emph{\method}, on the challenging V-COCO (\emph{Verbs in COCO}) dataset \cite{Gupta2015} for detecting human-object interactions. Our human-centric model improves accuracy by 26\% (relative) from 31.8 to 40.0 AP (evaluated by Average Precision on a triplet, called `role AP' \cite{Gupta2015}), with the gain mainly due to inferring the target object's relative position from the human appearance. In addition, we prove the effectiveness of \method by reporting a 27\% relative improvement on the newly released HICO-DET dataset \cite{hicodet}. Finally, our method can run at about 135ms / image for this complex task, showing good potential for practical usage.

\section{Related Work}\label{sec:related}

\paragraph{Object Detection.} Bounding-box based object detectors have improved steadily in the past few years. R-CNN, a particularly successful family of methods \cite{Girshick2014,Girshick2015,Ren2015}, is a two-stage approach in which the first stage proposes candidate RoIs and the second stage performs object classification. Region-wise features can be rapidly extracted \cite{He2014,Girshick2015} from shared feature maps by an RoI pooling operation. Feature sharing speeds up instance-level detection and enables recognizing higher-order interactions, which would be computationally infeasible otherwise. Our method is based on the Fast/Faster R-CNN frameworks \cite{Girshick2015,Ren2015}.

\paragraph{Human Action \& Pose Recognition.} The action and pose of humans is indicative of their interactions with objects or other people in the scene. There has been great progress in understanding human actions \cite{Gkioxari2015a} and poses \cite{Wei2016,Cao2017, he2017maskrcnn} from images. These methods focus on the human instances and do not predict interactions with other objects. We rely on action and pose appearance cues in order to predict the interactions with objects in the scene.

\paragraph{Visual Relationships.} Research on visual relationship modeling \cite{Sadeghi2011,Gupta2015,Lu2016,Yatskar2016} has attracted increasing attention. Recently, Lu \etal \cite{Lu2016} proposed to recognize visual relationships derived from an open-world vocabulary. The set of relationships include verbs (\eg, \cat{wear}), spatial (\eg, \cat{next to}), actions (\eg, \cat{ride}) or a preposition phrase (\eg, \cat{drive on}). Our focus is related, but different. First, we aim to understand \emph{human-centric} interactions, which take place in particularly diverse and interesting ways. These relationships involve direct interaction with objects (\eg, \cat{person cutting cake}), unlike spatial or prepositional phrases (\eg, \cat{dog next to dog}). Second, we aim to build detectors that recognize interactions in images with high precision, which is a requirement for practical applications. In contrast, in an open-world recognition setting, evaluating precision is not feasible, resulting in recall-based evaluation, as in \cite{Lu2016}.

\paragraph{Human-Object Interactions.} Human-object interactions \cite{Gupta2009,Yao2010,Delaitre2010} are related to visual relationships, but present different challenges. Human actions are more fine-grained (\eg, \cat{walking}, \cat{running}, \cat{surfing}, \cat{snowboarding}) than the actions of general subjects, and an individual person can simultaneously take multiple actions (\eg, \cat{drinking tea} and \cat{reading a newspaper} while \cat{sitting in a chair}). These issues require a deeper understanding of human actions and the objects around them and in much richer ways than just the presence of the objects in the vicinity of a person in an image. Accurate recognition of human-object interaction can benefit numerous tasks in computer vision, such as action-specific image retrieval \cite{Ramanathan2015}, caption generation \cite{Yu2015}, and question answering \cite{Yu2015, Mallya2016}.

\begin{figure}[t]
\centering
\includegraphics[width=1.\linewidth]{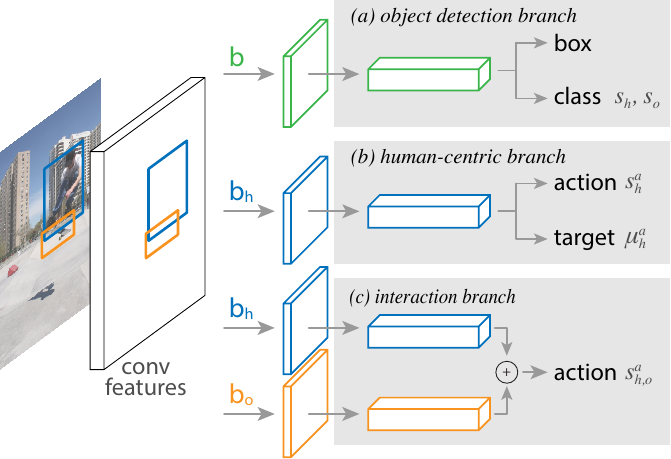}
\caption{\textbf{Model Architecture.} Our model consists of (a) an \emph{object detection} branch, (b) a \emph{human-centric} branch, and (c) an optional \emph{interaction} branch. The person features and their layers are shared between the human-centric and interaction branches (blue boxes).}
\label{fig:arch}
\end{figure}

\begin{figure*}[t]
\centering
\includegraphics[width=1.\linewidth]{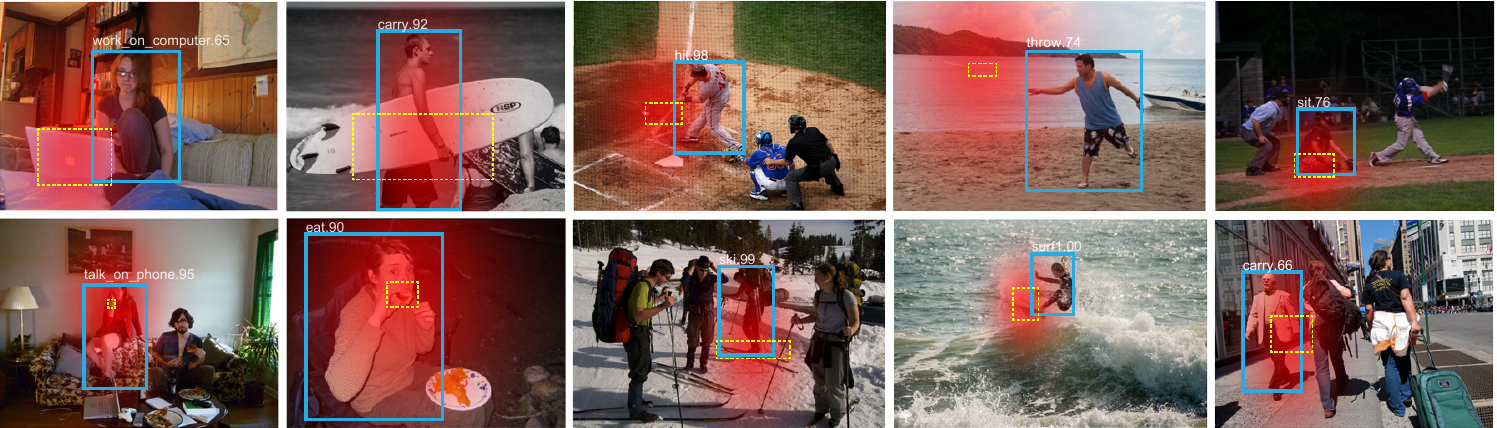}
\caption{\textbf{Estimating target object density from the person features}. We estimate a 4-d Gaussian density whose mean $\mu_h^a$ represents a 4-d offset for the target object of action $a$ (illustrated as yellow boxes); the variance of the density is illustrated in red for the 2-d translation offsets of $(x, y)$ (the scaling offsets' variance is not visualized). These target locations will be combined with the object detections $b_o$ to detect human-object interaction triplets. This figure also shows the predicted actions and their scores from the person RoIs. \textbf{The rightmost column} shows two intriguing examples: even though there are no target objects, our model predicts reasonably densities from the human pose (these predictions will be rejected by the object detection module, which will not find an object in the high density regions).}
\label{fig:target}
\end{figure*}

\section{Method}\label{sec:method}

We now describe our method for detecting human-object interactions. Our goal is to detect and recognize triplets of the form \hvo. To detect an interaction triplet, we have to accurately localize the box containing a \symb{human} and the box for the associated \symb{object} of interaction (denoted by $b_h$ and $b_o$, respectively), as well as identify the action $a$ being performed (selected from among $A$ actions).

Our proposed solution decomposes this complex and multifaceted problem into a simple and manageable form. We extend the Fast R-CNN \cite{Girshick2015} object detection framework with an additional \emph{human-centric} branch that classifies actions and estimates a probability density over the \emph{target} object location for each action. The human-centric branch reuses features extracted by Fast R-CNN for object detection so its marginal computation is lightweight.

Specifically, given a set of candidate boxes, Fast R-CNN outputs a set of object boxes and a class label for each box. Our model extends this by assigning a triplet score $S_{h,o}^a$ to pairs of candidate human/object boxes $b_h$, $b_o$ and an action $a$. To do so, we decompose the triplet score into four terms:
 \eqnnm{score}{S_{h,o}^a = s_h \cdot s_o \cdot s_h^a \cdot g_{h,o}^a }
While the model has multiple components, the basic idea is straightforward. $s_h$ and $s_o$ are the class scores from Fast R-CNN of $b_h$ and $b_o$ containing a human and object. Our human-centric branch outputs two extra terms. First, $s_h^a$ is the score assigned to action $a$ for the person at $b_h$. Second, $\mu_h^a$ is the predicted location of the target of interaction for a given human/action pair, computed based on the appearance of the human. This, in turn, is used to compute $g_{h,o}^a$, the likelihood that an object with box $b_o$ is the actual target of interaction. We give details shortly and show that this \emph{target localization} term is key for obtaining good results.

We discuss each component next, followed by an extension that replaces the action classification output $s_h^a$ with a dedicated \emph{interaction} branch that outputs a score $s_{h,o}^a$ for an action $a$ based on both the human and object appearances. Finally we give details for training and inference. Figure~\ref{fig:arch} illustrates each component in our full framework.

\subsection{Model Components}

\paragraph{Object Detection.} The object detection branch of our network, shown in Figure~\ref{fig:arch}(a), is identical to that of Faster R-CNN \cite{Ren2015}. First, a Region Proposal Network (RPN) is used to generate object proposals \cite{Ren2015}. Then, for each proposal box $b$, we extract features with RoiAlign \cite{he2017maskrcnn}, and perform object classification and bounding-box regression to obtain a new set of boxes, each of which has an associated score $s_o$ (or $s_h$ if the box is assigned to the person category). These new boxes are only used during inference; during training all branches are trained with RPN proposal boxes.

\paragraph{Action Classification.} The first role of the \emph{human-centric} branch is to assign an action classification score $s_h^a$ to each human box $b_h$ and action $a$. Just like in the object classification branch, we extract features from $b_h$ with RoiAlign and predict a score for each action $a$. Since a human can simultaneously perform multiple actions (\eg, \cat{sit} and \cat{drink}), our output layer consists of \emph{binary} sigmoid classifiers for \emph{multi-label} action classification (\ie the predicted action classes do not compete). The training objective is to minimize the binary cross entropy losses between the ground-truth action labels and the scores $s_h^a$ predicted by the model.

\paragraph{Target Localization.} The second role of the \emph{human-centric} branch is to predict the target object location based on a person's appearance (again represented as features pooled from $b_h$). However, predicting the \emph{precise} target object location based only on features from $b_h$ is challenging. Instead, our approach is to predict a \emph{density} over possible locations, and use this output together with the location of actual detected objects to precisely localize the target.

We model the density over the target object's location as a Gaussian function whose mean is predicted based on the human appearance and action being performed. Formally, the human-centric branch predicts $\mu_h^a$, the target object's 4-d mean location given the human box $b_h$ and action $a$. We then write our target localization term as:
 \eqnnm{target}{g_{h,o}^a = \exp(\lVert b_{o|h} - \mu_h^a \rVert^2 / 2\sigma^2)}
We can use $g$ to test the compatibility of an object box $b_o$ and the predicted target location $\mu_h^a$. In the above, $b_{o|h}$ is the encoding of $b_o$ in coordinates relative to $b_h$, that is:
 \eqnnm{bbox_reg}{b_{o|h}=\{\frac{x_o-x_h}{w_h},\frac{y_o-y_h}{h_h},\log\frac{w_o}{w_h},\log\frac{h_o}{h_h}\}}
This is a similar encoding as used in Fast R-CNN \cite{Girshick2015} for bounding box regression. However, in our case $b_h$ and $b_o$ are \emph{two different objects} and moreover $b_o$ is not necessarily near or of the same size as $b_h$. The training objective is to minimize the smooth $L_1$ loss \cite{Girshick2015} between $\mu_h^a$ and $b_{o|h}$, where $b_o$ is the location of the ground truth object for the interaction. We treat $\sigma$ as a hyperparameter that we empirically set to $\sigma=0.3$ using the validation set.

Figure~\ref{fig:target} visualizes the predicted distribution over the target object's location for example human/action pairs. As we can see, a \cat{carrying} appearance suggests an object in the person's hand, a \cat{throwing} appearance suggests an object in front of the person, and a \cat{sitting} appearance implies an object below the person. We note that the yellow dashed boxes depicting $\mu_h^a$ shown in Figure~\ref{fig:target} are inferred from $b_h$ and $a$ and did not have direct access to the objects.

Intuitively, our formulation is predicated on the hypothesis that the features computed from $b_h$ contain a strong signal pointing to the target of an action, even if that target object is outside of $b_h$. We argue that such `outside-the-box' regression is possible because the person's appearance provides a strong clue for the target location. Moreover, as this prediction is action-specific and instance-specific, our formulation is effective even though we model the target location using a uni-modal distribution. In Section~\ref{sec:experiments} we discuss a variant of our approach which allows us to handle conditionally multi-modal distributions and predict multiple targets for a single action.

\paragraph{Interaction\hspace{.5mm}Recognition.} Our human-centric model scores actions based on the human appearance. While effective, this does not take into account the appearance of the target object. To improve the discriminative power of our model, and to demonstrate the flexibility of our framework, we can replace $s_h^a$ in (\ref{eq:score}) with an \emph{interaction} branch that scores an action based on the the appearance of both the human and target object. We use $s_{h,o}^a$ to denote this alternative term.

The computation of $s_{h,o}^a$ reuses the computation from $s_h^a$ and additionally in parallel performs a similar computation based on features extracted from $b_o$. The outputs from the two action classification heads, which are $A$-dimensional vectors of logits, are summed and passed through a sigmoid activation to yield $A$ scores. This process is illustrated in Figure~\ref{fig:arch}(c). As before, the training objective is to minimize the binary cross entropy losses between the ground-truth action labels and the predicted action scores $s_{h,o}^a$.

\subsection{Multi-task Training}\label{sec:training}

We approach learning human-object interaction as a \emph{multi-task} learning problem: all three branches shown in Figure~\ref{fig:arch} are trained jointly. Our overall loss is the sum of all losses in our model including: (1) the classification and regression loss for the object detection branch, (2) the action classification and target localization loss for the human-centric branch, and (3) the action classification loss of the interaction branch. This is in contrast to our cascaded inference described in \S\ref{sec:inference}, where the output of the object detection branch is used as input for the human-centric branch.

We adopt image-centric training \cite{Girshick2015}. All losses are computed over both RPN proposal and ground truth boxes as in Faster R-CNN \cite{Ren2015}. As in \cite{Girshick2015}, we sample at most 64 boxes from each image for the object detection branch, with a ratio of 1:3 of positive to negative boxes. The human-centric branch is computed over at most 16 boxes $b_h$ that are associated with the human category (\ie, their IoU overlap with a ground-truth person box is $\ge0.5$). The loss for the interaction branch is only computed on positive example triplets (\ie, $\langle b_h, a, b_o \rangle$ must be associated with a ground truth interaction triplet). All loss terms have a weight of one, except the action classification term in the human-centric branch has a weight of two, which we found performs better.

\subsection{Cascaded Inference}\label{sec:inference}

At inference, our goal is to find high-scoring triplets according to $S_{h,o}^a$ in (\ref{eq:score}). While in principle this has $O(n^2)$ complexity as it requires scoring every pair of candidate boxes, we present a simple cascaded inference algorithm whose dominant computation has $O(n)$ complexity.

\paragraphi{Object Detection Branch}: We first detect all objects (including the \cat{person} class) in the image. We apply non-maximum suppression (NMS) with an IoU threshold of 0.3 \cite{Girshick2015} on boxes with scores higher than 0.05 (set conservatively to retain most objects). This step yields a new \emph{smaller} set of $n$ boxes $b$ with  scores $s_h$ and $s_o$. Unlike in training, these new boxes are used as input to the remaining two branches.

\paragraphi{Human-Centric Branch}: Next, we apply the human-centric branch to all detected objects that were classified as \cat{human}. For each action $a$ and detected human box $b_h$, we compute $s_h^a$, the score assigned to $a$, as well as $\mu_h^a$, the predicted mean offset of the target object location relative to $b_h$. This step has a complexity of $O(n)$.

\paragraphi{Interaction Branch}: If using the optional interaction branch, we must compute $s_{h,o}^a$ for each action $a$ and pair of boxes $b_h$ and $b_o$. To do so we first compute the logits for the two action classification heads independently for each box $b_h$ and $b_o$, which is $O(n)$. Then, to get scores $s_{h,o}^a$, these logits are summed and passed through a sigmoid for each pair. Although this last step is $O(n^2)$, in practice its computational time is negligible.

Once all individual terms have been computed, the computation of (\ref{eq:score}) is fast. However, rather than scoring every potential triplet, for each human/action pair we find the object box that maximizes $S_{h,o}^a$. That is we compute:
 \eqnnm{obj_pick}{b_{o^*}=\arg\max_{b_o}~~s_o \cdot s_{h,o}^a \cdot g_{h,o}^a}
Recall that $g_{h,o}^a$ is computed according to (\ref{eq:target}) and measures the compatibility between $b_o$ and the expected target location $\mu_h^a$. Intuitively, (\ref{eq:obj_pick}) encourages selecting a high-confidence object near the predicted target location of a high-scoring action. With $b_o$ selected for each $b_h$ and action $a$, we have a triplet of \hvo = $\langle b_h$, $a$, $b_o\rangle$. These triplets, along with the scores $S_{h,o}^a$, are the final outputs of our model. For actions that that do not interact with any object (\eg, \cat{smile}, \cat{run}), we rely on $s_h^a$ and the interaction output $s_{h,o}^a$ is not used, even if present. The score of such a predicted \hv pair is simply $s_h \cdot s_h^a$.

The above cascaded inference has a dominant complexity of $O(n)$, which involves extracting features for each of the $n$ boxes and forwarding through a small network. The pairwise $O(n^2)$ operations require negligible computation. In addition, for the entire system, a portion of computation is spent on computing the full-image shared convolutional feature maps. Altogether, our system takes \app135ms on a typical image running on a single Nvidia M40 GPU.

\begin{figure*}[t]
\centering\vspace{-1mm}
\includegraphics[width=1.\linewidth]{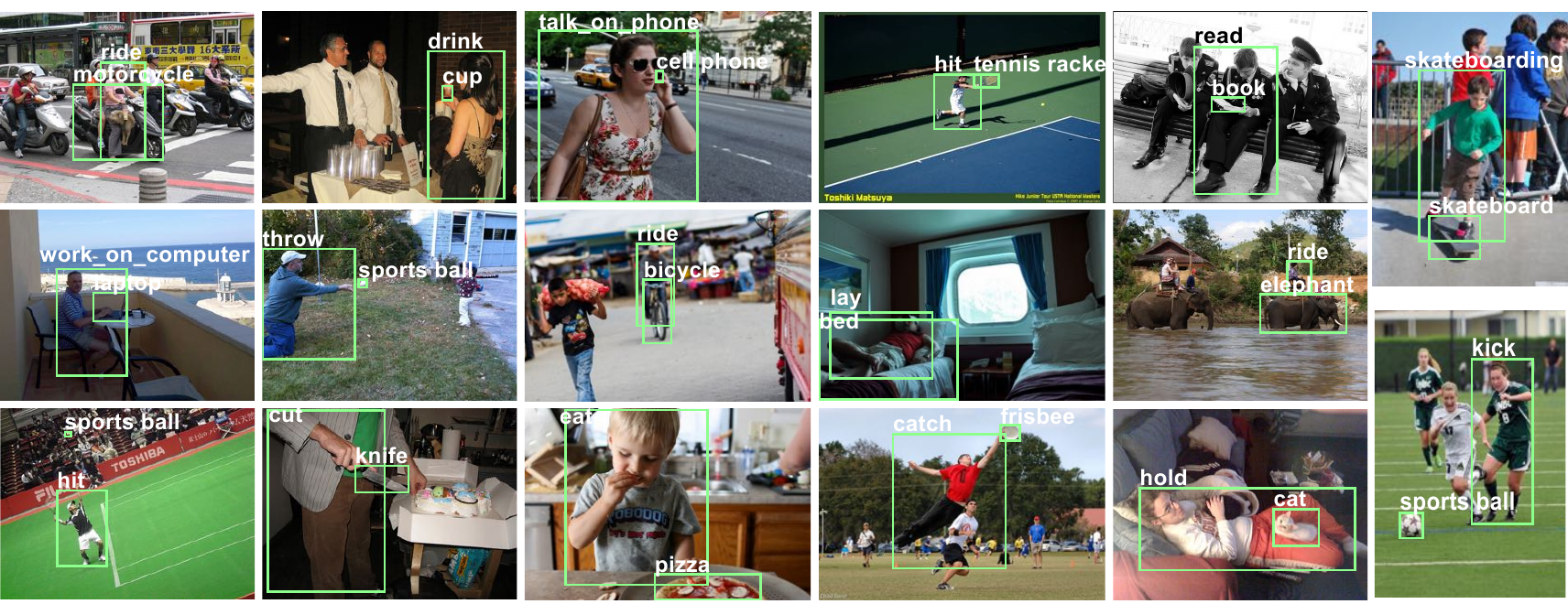}
\caption{Our results on some V-COCO test images. Each image shows one detected \hvo triplet.}
\label{fig:results_main}\vspace{-2mm}
\end{figure*}

\section{Datasets and Metrics}\label{sec:dataset}\vspace{-.25mm}

There exist a number of datasets for human-object interactions \cite{Le2014, Chao2015, Ronchi2015, Gupta2015, hicodet}. The most relevant for this work are V-COCO (\emph{Verbs in COCO}) \cite{Gupta2015} and HICO-DET~\cite{hicodet}. V-COCO serves as the primary testbed on which we demonstrate the effectiveness of \method and analyze its various components. The newly released HICO-DET~\cite{hicodet} contains \app48k images and 600 types of interactions and serves to further demonstrate the efficacy of our approach. The older TUHOI \cite{Le2014} and HICO \cite{Chao2015} datasets only have image-level labels and thus do not allow for grounding interactions in a \emph{detection} setting, while COCO-a \cite{Ronchi2015} is promising but only a small beta-version is currently available.

V-COCO is a subset of COCO \cite{Lin2014} and has \app5k images in the \texttt{trainval} set and \app5k images in the \texttt{test} set.\footnote{V-COCO's \texttt{trainval} set is a subset of COCO's \texttt{train} set, and its \texttt{test} set is a subset of COCO's \texttt{val} set. See \cite{Gupta2015} for more details. In this work, COCO's \texttt{val} images are not used during training in any way.} The \texttt{trainval} set includes \app8k person instances and on average 2.9 actions/person. V-COCO is annotated with 26 common action classes (listed in Table~\ref{table:res_vcoco}). Of note, there are three actions (\cat{cut}, \cat{hit}, \cat{eat}) that are annotated with two types of targets: \emph{instrument} and \emph{direct object}. For example, \cat{cut}+~\cat{knife} involves the instrument (meaning `cut with a knife'), and \cat{cut}+~\cat{cake} involves the direct object (meaning `cut a cake'). In \cite{Gupta2015}, accuracy is evaluated separately for the two types of targets. To address this, for the target estimation, we train and infer two types of targets for these three actions (\ie, they are treated like six actions for target estimation).

Following \cite{Gupta2015}, we evaluate two Average Precision (AP) metrics. We note that this is a \emph{detection} task, and both AP metrics measure \emph{both} recall and precision. This is in contrast to metrics of Recall@$N$ that ignore \emph{precision}.

The AP of central interest in the \emph{human-object interaction} task is the AP of the \emph{triplet} \hvo, called `\emph{role} AP' (AP$_\text{role}$) in \cite{Gupta2015}. Formally, a triplet is considered as a \emph{true positive} if: \textbf{(i)} the predicted human box $b_h$ has IoU of 0.5 or higher with the ground-truth human box, \textbf{(ii)} the predicted object box $b_o$ has IoU of 0.5 or higher with the ground-truth target object, \emph{and} \textbf{(iii)} the predicted and ground-truth actions match. With this definition of a true positive, the computation of AP is analogous to standard object detection (\eg, PASCAL \cite{Everingham2010}). Note that this metric does not consider the correctness of the target object category (but only the target object box location). Nevertheless, our method can predict the object categories, as shown in the visualized results (Figure~\ref{fig:results_simple} and Figure~\ref{fig:results_main}).

We also evaluate the AP of the pair \hv, called `\emph{agent} AP' (AP$_\text{agent}$) in \cite{Gupta2015}, computed using the above criteria of (i) and (iii). AP$_\text{agent}$ is applicable when the action has no object. We note that AP$_\text{agent}$ does note require localizing the target, and is thus of secondary interest.

\section{Experiments}\label{sec:experiments}

\paragraph{Implementation Details.} Our implementation is based on Faster R-CNN \cite{Ren2015} with a Feature Pyramid Network (FPN) \cite{Lin2017fpn} backbone built on ResNet-50 \cite{He2016}; we also evaluate a non-FPN version in ablation experiments. We train the Region Proposal Network (RPN) \cite{Ren2015} of Faster R-CNN following \cite{Lin2017fpn}. For convenient ablation, RPN is frozen and does not share features with our network (we note that feature sharing is possible \cite{Ren2015}). We extract 7$\times$7 features from regions by RoiAlign \cite{he2017maskrcnn}, and each of the three model branches (see Figure~\ref{fig:arch}) consist of two 1024-d fully-connected layers (with ReLU \cite{Nair2010}) followed by specific output layers for each output type (box, class, action, target).

Given a model pre-trained on ImageNet \cite{Deng2009}, we first train the object detection branch on the COCO \texttt{train} set (excluding the V-COCO \texttt{val} images). This model, which is in essence Faster R-CNN, has 33.8 object detection AP on the COCO \texttt{val} set. Our full model is initialized by this object detection network. We prototype our human-object interaction models on the V-COCO \texttt{train} split and perform hyperparameter selection on the V-COCO \texttt{val} split. After fixing these parameters, we train on V-COCO \texttt{trainval} (5k images) and report results on the 5k V-COCO \texttt{test} set.

We fine-tune our human-object interaction models for 10k iterations on the V-COCO \texttt{trainval} set with a learning rate of 0.001 and an additional 3k iterations with a rate of 0.0001. We use a weight decay of 0.0001 and a momentum of 0.9. We use synchronized SGD \cite{LeCun1989} on 8 GPUs, with each GPU hosting 2 images (so the effective mini-batch size per iteration is 16 images). The fine-tuning time is $\app$2.5 hours on the V-COCO \texttt{trainval} set on 8 GPUs.

\paragraph{Baselines.} To have a fair comparison with Gupta \& Malik \cite{Gupta2015}, which used VGG-16 \cite{Simonyan2015}, we reimplement their best-performing model (`model C' in \cite{Gupta2015}) using the same ResNet-50-FPN backbone as ours. In addition, \cite{Gupta2015} only reported AP$_\text{role}$ on a subset of 19 actions, but we are interested in \emph{all} actions (listed in Table~\ref{table:res_vcoco}). We therefore report comparisons in both the 19-action and all-action cases.

The baselines from \cite{Gupta2015} are shown in Table~\ref{table:res_baseline}. Our reimplementation of \cite{Gupta2015} is solid: it has 37.5 AP$_\text{role}$ on the 19 action classes tested on the \texttt{val} set, 11 points higher than the 26.4 reported in \cite{Gupta2015}. We believe that this is mainly due to ResNet-50 and FPN. This baseline model, when trained on the \texttt{trainval} set, has 31.8 AP$_\text{role}$ on all action classes tested on the \texttt{test} set. This is a strong baseline (31.8 AP$_\text{role}$) to which we will compare our method.

\begin{figure*}[t]
\centering
\includegraphics[width=1.\linewidth]{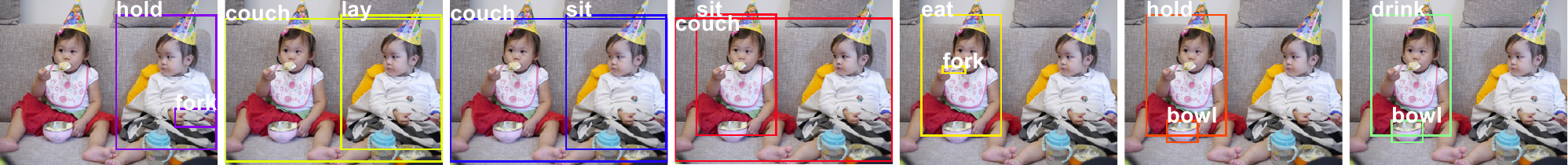}
\includegraphics[width=1.\linewidth]{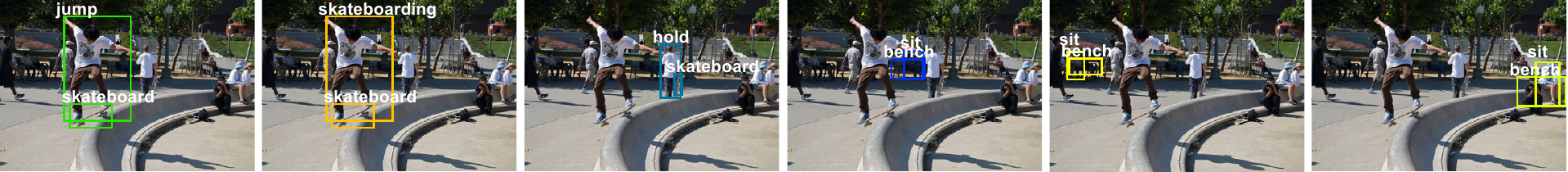}
\caption{\textbf{All detected triplets} on two V-COCO test images. We show all triplets whose scores (\ref{eq:score}) are higher than 0.01.}
\label{fig:results_multi_instances}\vspace{-3mm}
\end{figure*}

\begin{figure}[t]
\centering
\includegraphics[width=1.\linewidth]{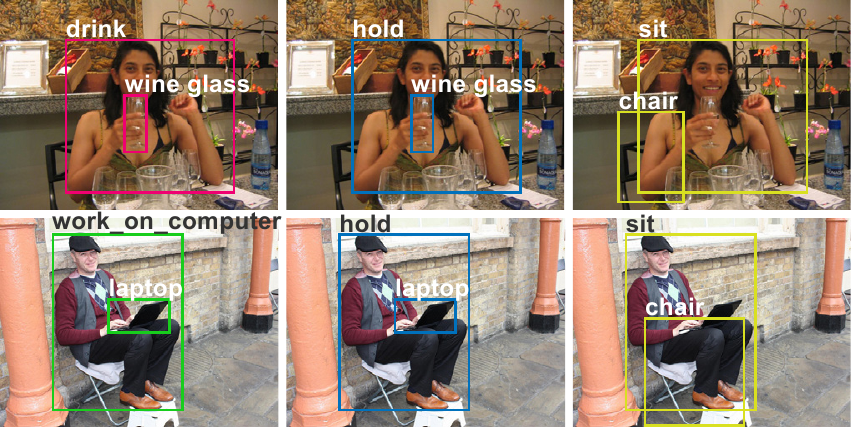}
\caption{Results of \method on test images. An individual person can take multiple actions and affect multiple objects.}
\label{fig:results_multi_actions}
\end{figure}

Our method, \method, has an AP$_\text{role}$ of \bd{40.0} evaluated on all action classes on the V-COCO \texttt{test} set. This is an absolute gain of 8.2 points over the strong baseline's 31.8, which is a relative improvement of 26\%. This result quantitatively shows the effectiveness of our approach.

\paragraph{Qualitative Results.} We show our human-object interaction detection results in Figure~\ref{fig:results_simple} and Figure~\ref{fig:results_main}. Each subplot illustrates one detected \hvo triplet, showing the location of the detected person, the action taken by this person, and the location (and category) of the detected target object for this person/action. Our method can successfully detect the object outside of the person bounding box and associate it to the person and action.

Figure~\ref{fig:results_multi_actions} shows our correctly detected triplets of one person taking multiple actions on multiple objects. We note that in this task, one person can take multiple actions and affect multiple objects. This is part of the ground-truth and evaluation and is unlike traditional object detection tasks \cite{Everingham2010} in which one object has only one ground-truth class.

Moreover, \method can detect multiple interaction instances in an image. Figure~\ref{fig:results_multi_instances} shows two test images with \emph{all detected triplets} shown. Our method detects multiple persons taking different actions on different target objects.

The multi-instance, multi-action, and multi-target results in Figure~\ref{fig:results_multi_instances} and Figure~\ref{fig:results_multi_actions} are all detected by one forward pass in our method, running at about 135ms per image on a GPU.

\begin{table}[t]
\tablestyle{5pt}{1.05}
\begin{tabular}{l|c|c}
 & 19 actions & all actions \\[-.5ex]
 mean AP$_\text{role}$ evaluated on
 & (on \texttt{val}) & (on \texttt{test}) \\
\shline
 model B of \cite{Gupta2015} [VGG-16] & 7.9 & N/A \\
 model C of \cite{Gupta2015} [VGG-16] & 26.4 & N/A \\
 model C of \cite{Gupta2015} [ResNet-50-FPN] \bd{ours} & 37.5 & 31.8 \\
\hline
\bd{\method} [ResNet-50-FPN] & \bd{41.9} & \bd{40.0}
\end{tabular}\vspace{.5em}
\caption{Comparisons with Gupta \& Malik \cite{Gupta2015}. To have an apples-to-apples comparisons, we reimplement \cite{Gupta2015}'s `model C' using ResNet-50-FPN. In addition, \cite{Gupta2015} reported AP$_\text{role}$ on a subset consisting of 19 actions, and only on the \texttt{val} set. As we evaluate on \emph{all} actions (more details in Table~\ref{table:res_vcoco}), for fair comparison, we also report the mean AP$_\text{role}$ on these 19 actions of \texttt{val}, with models trained on \texttt{train}. Our reimplemented baseline of \cite{Gupta2015} is solid, and \method is considerably better than this baseline.}
\label{table:res_baseline}
\end{table}

\paragraph{Ablation Studies.}

In Table~\ref{table:res_all_val}--\ref{table:res_arch_val} we evaluate the contributions of different factors in our system to the results.

\paragraphi{With \vs without target localization.} Target localization, performed by the human-centric branch, is the key component of our system. To evaluate its impact, we implement a variant \emph{without} target localization. Specifically, for each type of action, we perform k-means clustering on the offsets between the target RoIs and person RoIs (via cross-validation we found $k=2$ clusters performs best). This plays a role similar to density estimation, but is \emph{not} aware of the person appearance and thus is not instance-dependent. Aside from this, the variant is the same as our full approach.


Table~\ref{table:res_all_val} (a) \vs (c) shows that our target localization contributes significantly to AP$_\text{role}$. Removing it shows a degradation of 5.6 points from 37.5 to 31.9. This result shows the effectiveness of our target localization (see Figure~\ref{fig:target}). The per-category results are in Table~\ref{table:res_vcoco}.

\newcommand{\midd}[1]{\multirow{2}{*}{{#1}}}
\newcommand{\objinstr}{\begin{tabular}[l]{@{}l@{}}{\footnotesize ({\it object})} \\ {\footnotesize ({\it instrument})}\end{tabular}}
\begin{table}[t]\center
\resizebox{.9\linewidth}{!}{\tablestyle{2.2pt}{1.05}
\begin{tabular}{p{64pt}|x{24}x{24}|x{24}x{24}|x{24}x{24}}
 & \multicolumn{2}{c|}{baseline \cite{Gupta2015}} & \multicolumn{2}{c|}{\method}
 & \multicolumn{2}{c}{\midd{\bd{\method}}} \\
 & \multicolumn{2}{c|}{our impl.} & \multicolumn{2}{c|}{w/o target loc.} \\
\cline{2-7}
 & AP$_\text{agent}$ & \textbf{AP$_\text{role}$}
 & AP$_\text{agent}$ & \textbf{AP$_\text{role}$}
 & AP$_\text{agent}$ & \textbf{AP$_\text{role}$}\\
\shline
 carry & 62.2 & 8.1 & 63.9 & 14.4 & 64.8 & \bd{33.1} \\
 catch & 47.0 & 37.0 & 53.4 & 38.5 & 57.1 & \bd{42.5} \\
 drink & 11.9 & 18.1 & 37.5 & 25.4 & 46.0 & \bd{33.8} \\
 hold & 79.4 & 4.0 & 77.3 & 10.6 & 80.1 & \bd{26.4} \\
 jump & 75.5 & 40.6 & 75.6 & 39.3 & 74.7 & \bd{45.1} \\
 kick & 60.9 & 67.9 & 68.6 & \bd{70.6} & 77.5 & 69.4 \\
 lay & 50.1 & 17.8 & 51.1 & 18.6 & 47.6 & \bd{21.0} \\
 look & 68.8 & 2.8 & 61.0 & 2.7 & 59.4 & \bd{20.2} \\
 read & 34.9 & 23.3 & 43.2 & 22.0 & 41.6 & \bd{23.9} \\
 ride & 73.2 & \bd{55.3} & 76.2 & 55.0 & 74.2 & 55.2 \\
 sit & 76.8 & 15.6 & 75.6 & 15.1 & 76.1 & \bd{19.9} \\
 skateboard & 89.9 & 74.0 & 90.9 & 71.7 & 90.0 & \bd{75.5} \\
 ski & 84.0 & 29.7 & 83.9 & 28.2 & 84.7 & \bd{36.5} \\
 snowboard & 81.3 & 52.8 & 81.1 & 50.6 & 81.1 & \bd{63.9} \\
 surf & 94.6 & 50.2 & 94.5 & 50.3 & 93.5 & \bd{65.7} \\
 talk-on-phone & 63.3 & 23.0 & 74.7 & 23.8 & 82.0 & \bd{31.8} \\
 throw & 54.0 & 36.0 & 53.9 & 35.7 & 58.1 & \bd{40.4} \\
 work-on-computer & 70.2 & 46.1 & 72.6 & 46.9 & 75.7 & \bd{57.3} \\
\shline
\midd{cut \, \objinstr} & \midd{61.2} & 16.5 & \midd{69.1} & 17.7 & \midd{73.6} & \bd{23.0} \\
 & & 15.1 & & 19.5 & & \bd{36.4} \\
\midd{eat \, \objinstr} & \midd{75.6} & 26.5 & \midd{80.4} & 26.5 & \midd{79.6} & \bd{32.4} \\
 & & 2.7 & & \bd{2.9} & & 2.0 \\
\midd{hit \, \objinstr} & \midd{82.8} & 56.7 & \midd{83.9} & 55.3 & \midd{88.0} & \bd{62.3} \\
 & & 42.4 & & 41.3 & & \bd{43.3} \\
\shline
 point & 5.0 & -- & 4.0 & -- & 1.8 & -- \\
 run & 76.9 & -- & 77.8 & -- & 77.2 & -- \\
 smile & 60.6 & -- & 60.3 & -- & 62.5 & -- \\
 stand & 88.5 & -- & 88.3 & -- & 88.0 & -- \\
 walk & 63.9 & -- & 63.5 & -- & 65.4 & -- \\
\shline
 \bd{mean AP} & 65.1 & 31.8 & 67.8 & 32.6 & {69.2} & \bd{40.0} \\
\end{tabular}}\vspace{.5em}
\caption{Detailed results on V-COCO \texttt{test}. We show two main baselines and \method for each action. There are 26 actions defined in \cite{Gupta2015}, and because 3 actions (\cat{cut}, \cat{eat}, \cat{hit}) involve two types of target objects (instrument and direct object), there are 26+3 entries (more details in \S~\ref{sec:dataset}). We bold the leading entries on AP$_\text{role}$.}
\label{table:res_vcoco}\vspace{-2mm}
\end{table}

\paragraphi{With \vs without the interaction branch.} We also evaluate a variant of our method when removing the interaction branch. We can instead use the action prediction from the human-centric branch (see Figure~\ref{fig:arch}). 
Table~\ref{table:res_all_val} (b) \vs (c) shows that removing the interaction branch reduces AP$_\text{role}$ just slightly by 0.7 point. This again shows the main effectiveness of our system is from the target localization.

\paragraphi{With \vs without FPN.} Our model is a generic human-object detection framework and can support various network backbones. We recommend using the FPN \cite{Lin2017fpn} backbone, because it performs well for \emph{small objects} that are more common in human-object detection.


\begin{table}[t]
\tablestyle{5pt}{1.05}
\begin{tabular}{l|cc}
 & AP$_\text{agent}$ & \bd{AP$_\text{role}$} \\[-1ex]
 & {\tiny$\langle$\texttt{human}, \texttt{verb}$\rangle$} 
 & \bd{\tiny$\langle$\texttt{human}, \texttt{verb}, \texttt{object}$\rangle$} \\
\shline
 baseline \cite{Gupta2015} (our implementation) & 62.1 & 31.0  \\
\hline
 (a) \method w/o target localization & 65.1 & 31.9 \\
 (b) \method w/o interaction branch & 65.5 & 36.8 \\
 (c) \method & \bd{68.0} & \bd{37.5}
\end{tabular}\vspace{.5em}
\caption{Ablation studies on the V-COCO \texttt{val} set, evaluated by AP$_\text{agent}$ (\ie, AP of the \hv pairs)
and AP$_\text{role}$ (\ie, AP of the \hvo triplets). All methods are based on ResNet-50-FPN, including our reimplementation of \cite{Gupta2015}. Table~\ref{table:res_vcoco} shows the detail numbers of three entries: baseline, \method without target density estimation, and our complete method on the V-COCO \texttt{test} set. }
\label{table:res_all_val}
\end{table}


\begin{table}[t]
\tablestyle{5pt}{1.05}
\begin{tabular}{l|cc}
 & AP$_\text{agent}$ & \bd{AP$_\text{role}$} \\[-1ex]
 & {\tiny$\langle$\texttt{human}, \texttt{verb}$\rangle$} & \bd{\tiny$\langle$\texttt{human}, \texttt{verb}, \texttt{object}$\rangle$} \\
\shline
 \method w/ ResNet-50 & 65.0 & 35.9 \\
 \method w/ ResNet-50-FPN & \bf{68.0} & \bf{37.5} \\
\end{tabular}\vspace{.5em}
\caption{Ablation on the V-COCO \texttt{val} for vanilla ResNet-50 \vs ResNet-50-FPN \cite{Lin2017fpn} backbones.}
\label{table:res_fpn_val}
\end{table}


\begin{table}[t]
\tablestyle{2pt}{1.05}
\begin{tabular}{l|cc}
 & AP$_\text{agent}$ & \bd{AP$_\text{role}$} \\[-1ex]
 & {\tiny$\langle$\texttt{human}, \texttt{verb}$\rangle$} & \bd{\tiny$\langle$\texttt{human}, \texttt{verb}, \texttt{object}$\rangle$} \\
\shline
 \method w/ pairwise concat + MLP & 67.1 & \bd{37.5} \\
 \method & \bd{68.0} & \bd{37.5}
\end{tabular}\vspace{.5em}
\caption{Ablation on the V-COCO \texttt{val} set about the design of the pairwise interaction branch. See main text for explanations.}
\label{table:res_arch_val}
\end{table}


Table~\ref{table:res_fpn_val} shows a comparison between ResNet-50-FPN and a vanilla ResNet-50 backbone. The vanilla version follows the ResNet-based Faster R-CNN presented in \cite{He2016}. Specifically, the full-image convolutional feature maps are from the last residual block of the 4-th stage (res4), on which the RoI features are pooled. On the RoI features, each of the region-wise branches consists of the residual blocks of the 5-th stage (res5). Table~\ref{table:res_fpn_val} shows a degradation of 1.6 points in AP$_\text{role}$ when not using FPN. We argue that this is mainly caused by the degradation of the small objects' detection AP, as shown in \cite{Lin2017fpn}. Moreover, the vanilla ResNet-50 backbone is much slower, 225ms versus 135ms for FPN, due to use of res5 in the region-wise branches.


\paragraphi{Pairwise Sum \vs MLP.} In our interaction branch, the pairwise outputs from two RoIs are added (Figure~\ref{fig:arch}). Although simple, we have found that more complex variants do not improve results. We compare with a more complex transform in Table~\ref{table:res_arch_val}. We concatenate the two 1024-d features from the final fully-connected layers of the interaction branch for the two RoIs and feed it into an 2-layer MLP (512-d with ReLU for its hidden layer), followed by action classification. This variant is slightly worse (Table~\ref{table:res_arch_val}), indicating that it is not necessary to perform a complex pairwise transform (or there is insufficient data to learn this).

\begin{figure}[t]
\centering
\includegraphics[width=1.\linewidth]{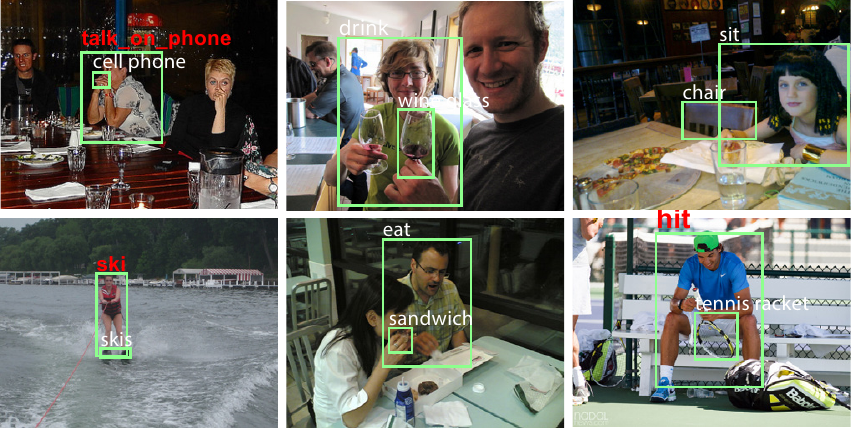}
\caption{False positive detections of our method.}
\label{fig:results_limitations}
\end{figure}

\begin{table}[t]
\tablestyle{5pt}{1.05}
\begin{tabular}{l|x{28}x{28}x{28}}
 method & full & rare & non-rare \\
\shline
 results from \cite{hicodet} & 7.81 & 5.37 & 8.54 \\
 baseline \cite{Gupta2015} (our impl.) & 9.09 & 7.02 & 9.71 \\
 \method & \bd{9.94} & \bd{7.16} & \bd{10.77}
\end{tabular}\vspace{.5em}
\caption{Results on HICO-DET \texttt{test} set. \method outperforms the approach in~\cite{hicodet} with a 27\% relative improvement. We also include our baseline approach, as described in Table~\ref{table:res_baseline}. }
\label{table:hicodet}\vspace{-3mm}
\end{table}

\paragraphi{Per-action accuracy.} Table~\ref{table:res_vcoco} shows the AP for each action category defined in V-COCO, for the baseline, \method without target localization, and our full system. We observe leading performance of AP$_\text{role}$ consistently. The actions with largest improvement are those with high variance in the spatial location of the object such as \cat{hold}, \cat{look}, \cat{carry}, and \cat{cut}. On the other hand, actions such as \cat{ride}, \cat{kick}, and \cat{read} show small or no improvement.

\paragraph{Failure Cases.} Figure~\ref{fig:results_limitations} shows some false positive detections. Our method can be incorrect because of false interaction inferences (\eg, top left), target objects of another person (\eg, top middle), irrelevant target objects (\eg, top right), or confusing actions (\eg, bottom left, \cat{ski} \vs \cat{surf}). Some of them are caused by a failure of \emph{reasoning}, which is an interesting open problem for future research.

\paragraph{Mixture Density Networks.} To improve target localization prediction, we tried to substitute the uni-modal regression network with a Mixture Density Network (MDN)~\cite{Bishop94mixturedensity}. The MDN predicts the mean and variance of $M$ relative locations for the objects of interaction conditioned on the human appearance. Note that MDN with $M=1$ is an extension of our original approach that also learns the variance in (\ref{eq:target}). However, we found that the MDN layer does not improve accuracy. More details and discussion regarding the MDN experiments can be found in Appendix A.

\paragraph{HICO-DET Dataset.} We additionally evaluate \method on HICO-DET \cite{hicodet} which contains 600 types of interactions, composed of 117 unique verbs and 80 object types (identical to COCO objects). We train \method on the \texttt{train} set, as specified by the authors, and evaluate performance on the \texttt{test} set using released evaluation code. Results are shown in Table~\ref{table:hicodet} and discussed more in Appendix B.

\section*{Appendix A: Mixture Density Networks}\label{appendix:a}

The target localization module in \method learns to predict the location of the object of interaction conditioned on the appearance of the human hypothesis $h$. As an alternative, we can allow the relative location of the object to follow a multi-modal conditional distribution. For this, we replace our target localization module with a Mixture Density Network (MDN)~\cite{Bishop94mixturedensity}, which parametrizes the mean, variance and mixing coefficients of $M$ components of the conditional normal distribution ($M$ is a hyperparameter).  This instantiation of \method is flexible can can capture different modes for the location of the objects of interaction. 

The localization term for scoring $b_{o|h}$ is defined as:
\eqnnm{targetm}{g_{h,o}^a = \sum_{m=0}^{M-1} w^{a,m}_h \cdot \mathcal{N} \bigl( b_{o|h} \hspace{1mm} |  \hspace{1mm} \mu^{a,m}_h, \sigma^{a,m}_h \bigr) }
The mixing coefficients are required to have $w^{m}_h \in [0, 1]$ and $\sum_m w^{a,m}_h = 1$. We parametrize $\mu$ as a 4-D vector for each component and action type. We assume a diagonal covariance matrix and thus parametrize $\sigma$ as a 4-D vector for each component and action type. Compare the localization term in (\ref{eq:targetm}) with the term in (\ref{eq:target}). Inference is unchanged, except we use the new form of $g_{h,o}^a$ when evaluating (\ref{eq:obj_pick}).

To instantiate MDN, the human-centric branch of our network must predict $w$, $\mu$, and $\sigma$. We train the network to minimize $-\log(g_{h,o}^a)$ given $b_o$ (the location of the ground truth object for each interaction). We use a fully connected layer followed by a softmax operation to predict the mixing coefficients $w^{a,m}_h$. For $\mu$ and $\sigma$, we also use fully connected layers. However, in the case of $\sigma$, we use a softplus operation ($f(x) = \log(e^x +1)$) to enforce positive values for the covariance coefficients. This leads to stable training, compared to a variant which parametrizes $\log\sigma^2$ and becomes unstable due to an exponential term in the gradients. In addition, we found that enforcing a lower bound on the covariance coefficients was necessary to avoid overfitting. We set this value to 0.3 throughout our experiments.

Table~\ref{table:mixture} shows the performance of \method with MDN and compare it to our original model. Note that the MDN with $M=1$ is an extension of our original approach, with the only difference that it learns the variance in (\ref{eq:target}). The performance of MDN with $M=1$ is similar to our original model, which suggests that a learned variance does not lead to an improvement. With $M=2$, we do not see any gains in performance, possibly because the human appearance is a strong cue for predicting the object's location and possibly due to the limited number of objects per action type in the dataset. Based on these results, along with the relative complexity of MDN, we chose to use our simpler proposed target localization model instead of MDN. Nonetheless, our experiments demonstrate that MDN can be trained within the \method framework, which may prove useful.

\begin{table}[t]
\tablestyle{5pt}{1.05}
\begin{tabular}{l|cc}
 & AP$_\text{agent}$ & \bd{AP$_\text{role}$} \\[-1ex]
 & {\tiny$\langle$\texttt{human}, \texttt{verb}$\rangle$} 
 & \bd{\tiny$\langle$\texttt{human}, \texttt{verb}, \texttt{object}$\rangle$} \\
\shline
 \method  & 68.0 & 37.5 \\
 \method w/ MDN ($M=1$) & 68.2 & 37.7 \\
 \method w/ MDN ($M=2$) & 68.0 & 37.3
\end{tabular}\vspace{.5em}
\caption{Ablation studies for MDN on the V-COCO \texttt{val} set, evaluated by AP$_\text{agent}$ (\ie, AP of the \hv pairs)
and AP$_\text{role}$ (\ie, AP of the \hvo triplets). All methods are based on ResNet-50-FPN.}
\label{table:mixture}
\end{table}

\begin{figure}[t!]
\centering
\includegraphics[width=1.\linewidth]{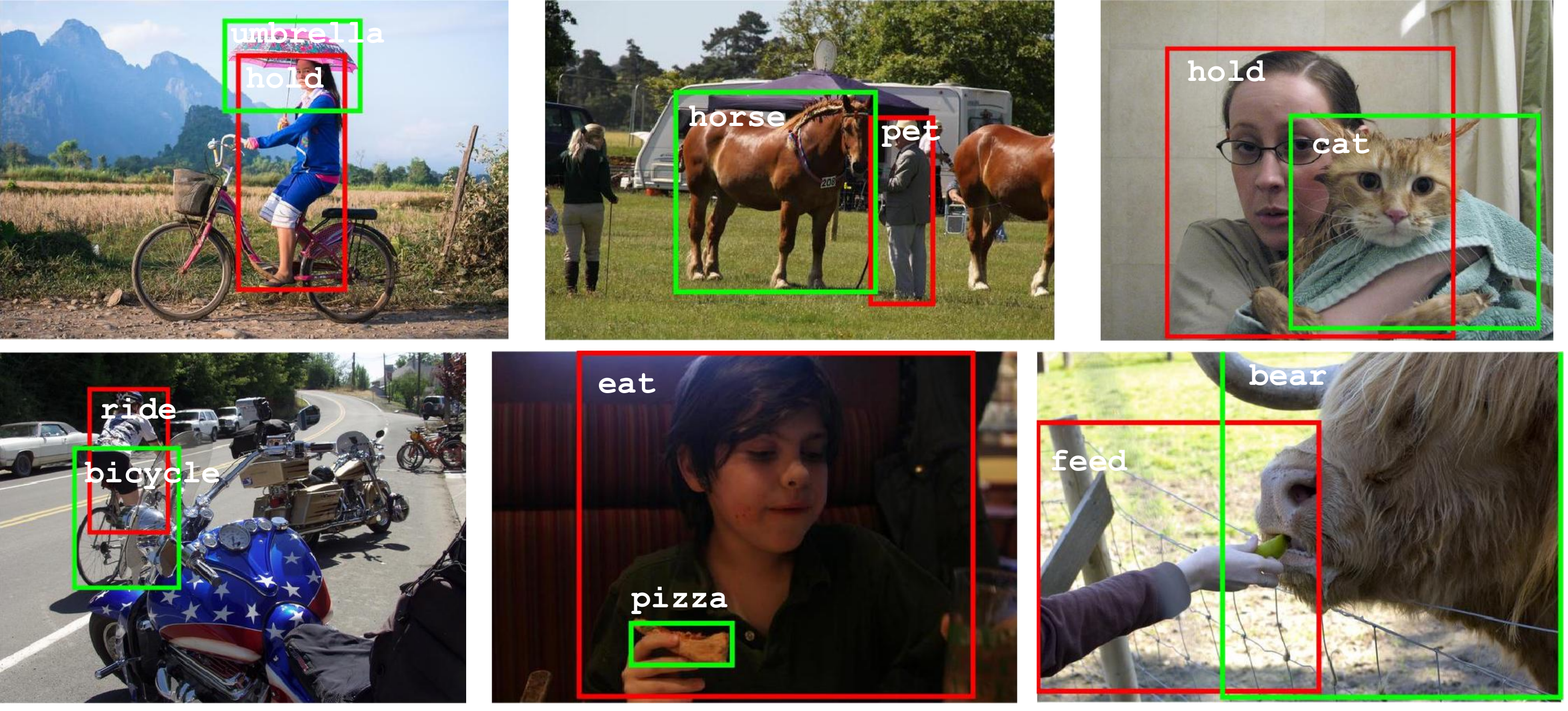}
\caption{Our results on the HICO-DET \texttt{test} set. Each image shows one detected \hvo triplet. The red box denotes the detected human instance, while the green box the detected object of interaction.}
\label{fig:fighico}
\end{figure}

\section*{Appendix B: HICO-DET Dataset}\label{appendix:b}

As discussed, we also test \method on the HICO-DET dataset~\cite{hicodet}. HICO-DET contains approximately 48k images and is annotated with 600 interaction types. The annotations include boxes for the humans and the objects of interactions. There are 80 unique object types, identical to the COCO object categories, and 117 unique verbs.

Objects are not exhaustively annotated on HICO-DET. To address this, we first detect objects using a ResNet50-FPN object detector trained on COCO as described in~\cite{Lin2017fpn}. These detections are kept frozen during training, by setting a zero-valued weight on the object detection loss of \method. To train the human-centric and interaction branch, we assign ground truth labels from the HICO-DET annotations to each person and object hypothesis based on box overlap. To diversify training, we jitter the object detections to simulate a set of region proposals. We define 117 action types and use the object detector to identify the type for the object of interaction, \eg orange vs. apple.

We train \method on the \texttt{train} set defined in \cite{hicodet}, for 80k iterations and with a learning rate of 0.001 (a 10x step decay is applied after 60k iterations). In the interaction branch, we use dropout with a ratio of 0.5. We evaluate performance on the \texttt{test} set using released evaluation code.

Results are shown in Table~\ref{table:hicodet}. We show a 27\% relative performance gain compared to the published results in~\cite{hicodet} and a 9\% relative gain compared to our baseline approach. Figure~\ref{fig:fighico} shows example predictions made by \method.

{\small\bibliographystyle{ieee}\bibliography{refs_extended}}

\end{document}